# A Miniature 3-DoF Flexible Parallel Robotic Wrist Using NiTi Wires for Gastrointestinal Endoscopic Surgery

Huxin Gao[1,2,3,†], Xiao xiao[4,†], Xiaoxiao Yang[5],
Tao Zhang[2], Xiuli Zuo[5], Yanqing Li[5], Hongliang Ren[2,1,3]

*Abstract*— Gastrointestinal endoscopic surgery (GES) has high requirements for instruments' size and distal dexterity, because of the narrow endoscopic channel and long, tortuous human gastrointestinal tract. This paper utilized Nickel-Titanium (NiTi) wires to develop a miniature 3-DoF (pitch-yaw-translation) flexible parallel robotic wrist (FPRW). Additionally, we assembled an electric knife on the wrist's connection interface and then teleoperated it to perform an endoscopic submucosal dissection (ESD) on porcine stomachs. The effective performance in each ESD workflow proves that the designed FPRW has sufficient workspace, high distal dexterity, and high positioning accuracy.

## I. INTRODUCTION

Gastrointestinal endoscopic surgery (GES) requires long and flexible instruments to pass through a narrow and tortuous endoscopic working channel (e.g., Olympus's products: small inner diameter: 2.0mm–2.8mm, length: 1200mm) compared with rigid instruments (e.g., Endowrist™ in da Vinci system) in laparoscopic surgery. Therefore, there is a size limitation and flexibility for GES instruments. Additionally, the instruments are required to perform a series of surgical procedures. Taking endoscopic submucosal dissection (ESD) as an example, the electric knife needs to execute marking, circumferential cutting and submucosal dissection, which has high distal dexterity requirements for GES instruments. However, current commercial ESD instruments have limited degrees of freedom (DoFs). For example, conventional electric knives and injection needles have only translational motion, so the distal bending motions depend on the endoscopic movement. Therefore, it is necessary to explore GES instruments with high distal dexterity. Finally, the emerging robot-assisted surgery remarkably benefits surgeons' performance and patients' recovery, typically laparoscopic surgery using the da Vinci system. Therefore, ESD instruments tend to be miniature, flexible, dexterous, and robot-based in the future.

This work is supported by the National Key R&D Program of China 2018YFB1307700, (including the associated subprogram 2018YFB1307703), The Ministry of Science and Technology (MOST) of China, and the National Natural Science Foundation of China (No.62073043, No.62003045). (Corresponding author: Hongliang Ren; email: hlren@ieee.org).

[1]Department of Biomedical Engineering, National University of Singapore, Singapore 117575, Singapore
[2]Department of Electronic Engineering, The Chinese University of Hong Kong (CUHK), Hong Kong.
[3]NUS (Suzhou) Research Institute (NUSRI), Suzhou 215123, China.
[4]Department of Electrical and Electronic Engineering, Southern University of Science and Technology, Shenzhen, China.
† Equal contribution.

The main method to increase distal dexterity is to utilize a multi-DoF wrist. Various robotic wrists have been developed as surgical instruments in the past decade. Typical forms of wrists include the serial type and parallel type. The serial linkage structure can form 2- or 3-DoF wrist because of high rigidity and positioning accuracy. Similar to the structure of Endowrist™, [1] proposed a 3-DoF (pitch-yaw-roll) wrist with cable guide channel for compact articulation. Moreover, gear structure [2], [3] was introduced to create a articulated joint for two rigid links to enhance the stiffness. However, a serial linkage-based wrist has a finite offset between the rotation axes, resulting in inconsistent coordination compared with a human wrist. To solve this problem, different kinds of bending mechanism with 2 DoFs (pitch-yaw) were developed, such as cable-driven sphere type [4], snake-bone type [5], flexible tube (e.g. hypotube) [6]. However, the large radius of curvature reduces the positioning accuracy and stiffness. Additionally, due to the close radial distance between the endoscope axis and instrument axis, the end-effector of serial instruments tends to be far away from surgeons' view, which will make surgeons uncomfortable in operation.

The parallel structure effectively reduces the wrist's size and improves the end-effector's positioning accuracy and stiffness. Parallel wrists can be developed based on rigid-link chains. [7] used three prismatic–spherical–revolute link chains to develop a 3-DoF (pitch-yaw-roll) parallel wrist with high stiffness and dexterity and another one grasping DoF. Likewise, [8] used three prismatic-universal chains to develop a wrist with variable stiffness. However, it is extremely difficult for the rigid links to form a miniature wrist with less than 3.0mm diameter to enter the gastroendoscope. Besides, origami structure was used to create a 3-DoF (pitch-yaw-roll) wrist [9] and integrate a grasper, but its size was also too big for GES.

To overcome the size limitation in the parallel wrist, in this paper, we develop a miniature 3-DoF flexible parallel robotic wrist (FPRW) using super-elastic Nickel-Titanium (NiTi) wires for GES. The following sections mainly introduce the robotic design and ex vivo test.

## II. ROBOTIC SYSTEM

In this work, the adopted endoscope (HUACO) has a narrow channel with a total length of 1200mm and an inner diameter of 2.8mm. According to these, we design a miniature NiTi-driven flexible parallel robotic wrist (FPRW) as shown in Fig. 1. The transmission consists of four internal tendon-sheaths, an outer PTFE tube and a wirerope of an electric knife through an inner PTFE tube. Its total length and outer diameter are 1380mm and 2.5mm. The following section mainly introduces the NiTi-driven wrist.

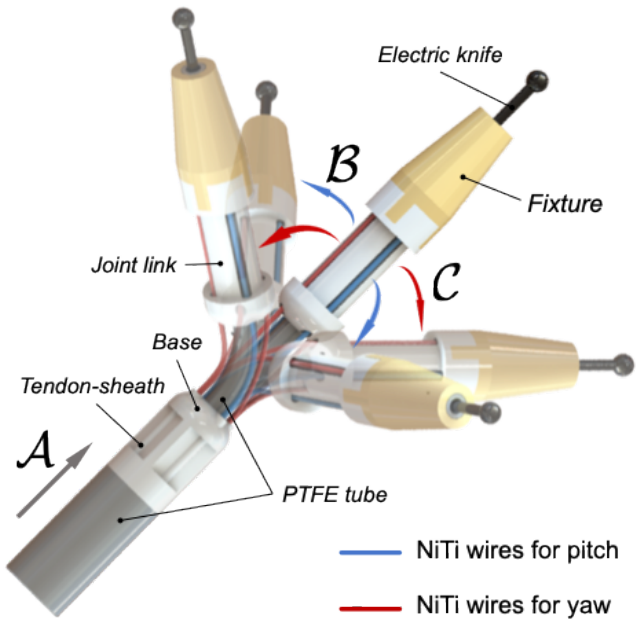

Fig. 1. FPRW's 3D model and motion demonstration.

The wrist, with a length of 20mm and a diameter of 2.5 mm, consists of a base, a joint link assembled with an electric knife as an end-effector. It has three DOFs as shown in Fig. 1: translation $A$ pitch $B$ and yaw $C$ DoF $A$ is driven by the PTFE tube. DoF $B$ & $C$ are formed by a NiTi-driven bending joint with a parallel structure and each one is driven by two NiTi cables. The end-effector is an insulating PEEK (a plastic material) fixture mounted with an electric knife, fixed on the joint link's connection interface. The electric knife's metal cable passes through an inner PTFE tube to insulate itself from the metal links. Besides, the distance between the base and joint link is kept in the range of 4 to 5mm to trade off the DoF $B$ & $C$ displacement and the wrist's stiffness. Additionally, the inner PTFE tube can increase the bending joint's stiffness.

## III. EX VIVO TEST

To validate the FPRW's performance in practice, an ex vivo test was conducted to perform an ESD procedure, as shown in Fig. 2A. The endoscope was inserted into the stomach through the esophagus and was adjusted manually to orient the lesion. Then, the FPRW was inserted into the endoscopic working channel with a diameter of 2.8mm. During ESD, the FPRW was firstly teleoperated via a master device (Touch, Geomagic) to make some discrete markers surrounding the lesion. Secondly, an injection needle was inserted into the other endoscopic working channel with a diameter of 3.7mm and then manually injected a mixture of normal saline and methylene blue into the submucosal layer to form a bump at the lesion area. Thirdly, the FPRW was controlled to cut a circumferential incision surrounding the markers, proving that the FPRW contains enough DOFs to follow a circular trajectory and stiffness to cut a thick mucosa. Fourthly, after the injection needle was withdrawn, an in-house manipulator with a diameter of 3.5mm was inserted into the 3.7mm working channel and was controlled to grasp and lift the edge of the mucosa surrounding the artificial lesion through the incision. Thanks to the open submucosal view and high FPRW's distal dexterity, the FPRW was easily teleoperated to dissect the submucosal layer until the tissue was cut off. This shows that the FPRW has a sufficient workspace to cover a lesion with a diameter of around 20mm. From Fig. 2B left, we find that few muscular injuries are observed on the dissection plane of the opened stomach, which indicates the safety of the FPRW during ESD operation. En bloc resection is achieved with all of the markers included in the resected tissue, as shown in Fig. 3B right, even though slight mucosal damage occurs. Additionally, no piecemeal resection is recorded in all of the ESD procedures.

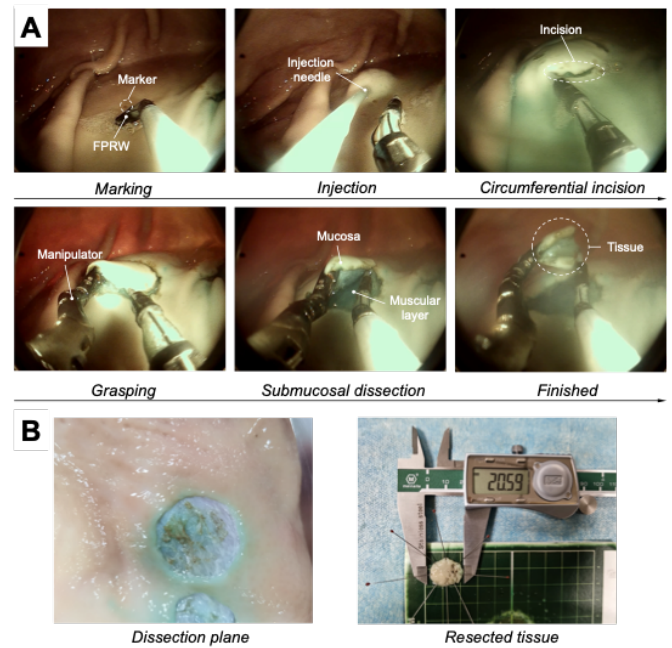

Fig. 2. Ex vivo test. **A**. ESD procedures. Marking – injection – circumferential incision – grasping (by left manipulator) – submucosal dissection – Finished. **B** ESD result. Left: dissection plane. Right: ressected tissue.

## IV. CONCLUSIONS AND FUTURE WORKS

This paper proposes a miniature 3-DoF flexible parallel robotic wrist (FPRW) based on NiTi wires. Additionally, we used the FPRW to perform a complete ESD which validates the FPRW's performance.

We need to analyze the wrist's stiffness in future work and then modify the design further. More surgical tools can be also integrated with the FPRW, such as biopsy forceps and injection needles.